\documentclass[10pt,twocolumn,letterpaper]{article}
\usepackage[affil-it]{authblk}
\usepackage{cvpr}              

\usepackage{graphicx}
\usepackage{amsmath}
\usepackage{amssymb}
\usepackage{booktabs}
\usepackage{algorithm}
\usepackage{algpseudocode}
\usepackage{multirow}
\usepackage{paralist}
\usepackage[table]{xcolor}
\usepackage{color,colortab}

\definecolor{lightgray}{gray}{0.9}
\definecolor{LightCyan}{rgb}{0.88,1,1}
\floatname{algorithm}{Procedure}

\def\@IEEEfigurecaptionsepspace{\vskip\abovecaptionskip\relax}%
\def\@IEEEtablecaptionsepspace{\vskip\abovecaptionskip\relax}%
\usepackage[pagebackref,breaklinks=true,colorlinks,citecolor=blue,urlcolor=blue,linkcolor=blue,bookmarks=false]{hyperref}

\usepackage[capitalize]{cleveref}
\crefname{section}{Sec.}{Secs.}
\Crefname{section}{Section}{Sections}
\Crefname{table}{Table}{Tables}
\crefname{table}{Tab.}{Tabs.}

\def\cvprPaperID{****} 
\def\confName{****}
\def\confYear{****}

\begin{document}

\title{Consistency-Aware Anchor Pyramid Network for Crowd Localization}

\author{
 \textit{Xinyan Liu$^{1}$ \quad Guorong Li$^{1}$\thanks{Corresponding author.} \quad Yuankai Qi$^{2}$} \\\quad Zhenjun Han$^{1}$ \quad Qingming Huang$^{1,3,4}$ \quad Ming-Hsuan Yang$^{5}$ \quad Nicu Sebe$^6$\\
$^{1}$ \small University of Chinese Academy of Science, Beijing, China \\
$^{2}$Australian Institute for Machine Learning, The University of Adelaide \\
$^{3}$Key Lab of Intell. Info. Process., Inst. of Comput. Tech., CAS, Beijing, China \\
$^{4}$Peng Cheng Laboratory, Shenzhen, China, $^{5}$University of California, Merced \\
$^{6}$University of Trento, Italy \\
{\small \{liuxinyan19@mails.ucas.ac.cn, \{liguorong, hanzhj, qmhuang\}@ucas.ac.cn, qykshr@gmail.com, mhyang@ucmerced.edu, sebe@disi.unitn.it\}}
}


\maketitle

\begin{abstract}
Crowd localization aims to predict the spatial position of humans in a crowd scenario.
We observe that the performance of existing methods is challenged from two aspects: (i) ranking inconsistency between test and training phases; and (ii) fixed anchor resolution may underfit or overfit crowd densities of local regions.
To address these problems, we design a supervision target reassignment strategy for training to reduce ranking inconsistency and 
propose an anchor pyramid scheme to adaptively determine the anchor density in each image region.  
Extensive experimental results on three widely adopted datasets (ShanghaiTech A\&B, JHU-CROWD++, UCF-QNRF) demonstrate the favorable performance against several state-of-the-art methods.
\end{abstract}

\section{Introduction}

\label{sec:intro}

\begin{figure}[!htb]
    \centering
\includegraphics[width=0.45\textwidth]{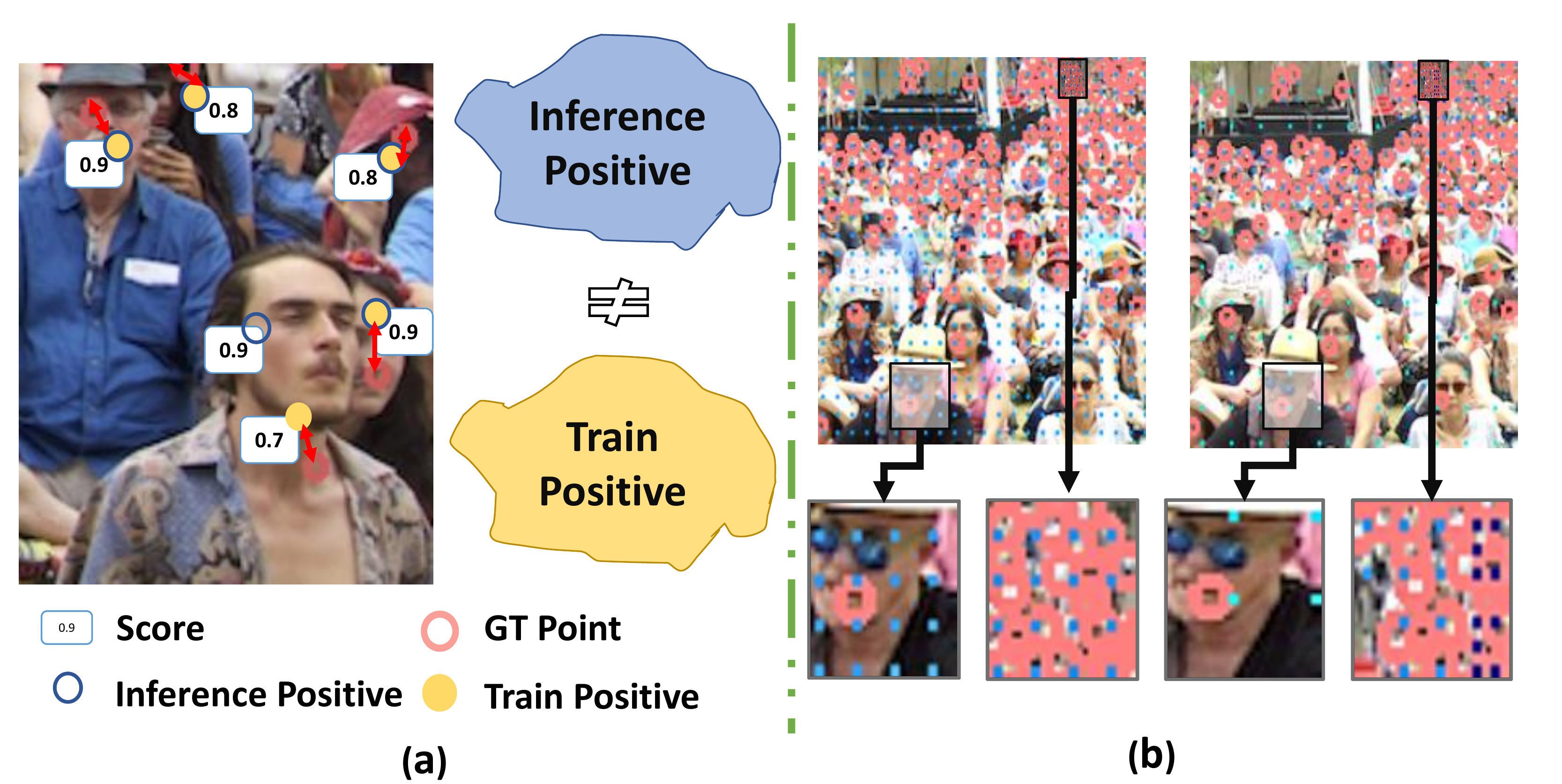}
\caption{Existing methods use a fixed number of evenly distributed anchors on all images of a dataset, which leads to redundant or insufficient target sparse or dense regions. We propose a pyramid anchor strategy to alleviate this issue.
}
\label{fig:anchor}
\end{figure} 

Crowd counting and localization aim to learn to count and localize crowds with only point annotations. It has received increasing attention due to its broad applications, such as traffic flow analysis and crowd abnormality detection. Despite recent advances ~\cite{MCNN,ppnet,ucfcomposition,wang2021self_training,context,csrnet,cross,countingbyflows,idrees2018composition,liang2021focal,wang2021dense,ma2019bayesian} in this field, it remains challenging due to the significant density variations in crowded scenes.

Crowd localization is usually formulated as a regression problem. Existing methods fall into two main groups according to their regression target: density map regression and coordinate regression.
The former predicts a density map from an input image and then compares against ground truth density maps generated by smoothing Gaussian Kernels~\cite{ma2019bayesian,wan2021generalized,wang2020distribution,kim2020probabilistic,liang2021focal}.
The local maximum pixels larger than a pre-defined threshold are considered as predicted locations.
Since the predicted density map is mainly optimized for counting, it is usually too smooth to generate accurate locations.

Following the paradigm of anchor-based object detection~\cite{yolo9000,redmon2018yolov3,ren2015faster,lin2017focal,redmon2016you,liu2016ssd}, coordinate regression methods utilize dense anchors to predict the coordinates of objects or humans directly.
Each anchor predicts a target coordinate offset and a probability of being a target.

Although significant progress has been made, existing coordinate regression methods suffer two limitations.

First, there is a ranking discrepancy during training and test. During training, only top-K (K is the number of targets in a training image) anchors are selected for gradient back-propagation, and the selection criteria are based on both localization distance and classification confidence. However, the top-K anchors during the test are selected merely based on classification, as shown in Figure.~\ref{fig:anchor}(a). 
Such selection discrepancy may lead to some anchors that should but are not used for network optimization.
Second, as shown in Figure.~\ref{fig:anchor}(b),
existing methods adopt a fixed number of evenly distributed anchors on all images in a dataset, which can be redundant in sparse target regions and insufficient in dense target regions.

To address these problems, we propose a consistency-aware anchor pyramid network for crowd localization.  
It consists of two main components:   Consistency-aware Target Rearrangement (CTR) and Adaptive Anchor Pyramid Selection (AAPS).
Specifically, the CTR module first selects two groups of top-K anchors according to separate criteria: the first group is chosen according to both conventional localization distance and classification confidence as existing methods, and each of them is assigned a ground truth annotation; the second group is chosen according to just classification confidence to keep consistency with the test phase. Then we reassign the first group's targets to the second group's anchors according to location distance, and the union of these two groups of anchors is used for gradient back-propagation. In such a way, the ranking discrepancy problem is alleviated.

Regarding the AAPS, we divide the input image into several grids. Multiple anchor densities are used in each grid to predict target locations. At the same time, we use another branch to predict the target number in each grid. Then, in each grid,  only the anchors belonging to a density that is most close to the predicted target number are retained. Next, all the retained anchors are used as candidates for computing the best matching with ground truth annotations. In this way, we not only fulfill using different anchor densities at different image regions but also vastly reduce the computation cost compared to directly using all anchors for best-matching computation.
The main contributions of the paper are: 
\begin{compactitem}
    \item We propose a consistency-aware target rearrangement (CTR) strategy for anchor-based crowd-counting methods, alleviating the ranking discrepancy problem during training and testing.
    \item We propose an adaptive anchor pyramid selection (AAPS) strategy to adaptively determine the anchor density in each image region with only a slight computation increase.
  \item Extensive experiments on the ShanghaiTech, JHU-CROWD++, and UCF-QNRF benchmark datasets demonstrate the effectiveness of the proposed method compared against several state-of-the-art approaches. 
\end{compactitem}

\section{Related Work}

There are two main types of crowd localization methods: density map regression and coordinate regression.

\noindent\textbf{Density map regression.}
A density map is generated by applying smoothing kernels on the point annotations. Existing density map regression methods mainly focus on better extracting coordinates methods and generating clear density maps.
Rebera \etal \cite{LOWB} propose a hand-craft method to extract coordinates from density maps. 
It involves selecting pixels whose confidence of being a target point is higher than around pixels (peak density point), filtering out low confidence points by a pre-defined threshold, and lastly, using clustering methods, such as K-means, on the filtered points to generate cluster center points. Those clustered center points are considered as locations of objects. 
Haroon~\cite{ucfcomposition} \etal use cascade adaptive Gaussian kernels to refine the density map to localization map. To extract coordinates from the localization map, they apply a Dirac Delta function to pixels with positive confidence. 
However, those post-refine methods assume that predicted density maps have clear peak points and no noise, which is hard to guarantee in practice. 
The method based on Bayesian loss~\cite{ma2019bayesian} learns from the point annotation by considering the maximum posterior probability. 
However, it still requires a Gaussian kernel function to calculate the conditional probability of different distances. 
The estimation of Gaussian kernel size depends on the object's density, and its accuracy will drop at the sparse region. DM-Count~\cite{wang2020distribution} and D2CNet~\cite{dcnet} decouple the counting 
task into two targets, total count learning, and normalized probability map learning. DM-Count uses the optimal transport method to measure the difference between the predicted density map and the point annotation map. The D2CNet estimates possible object boundaries to generate a probability map. Although it has better peak points for locating individual objects, the probability maps need to be normalized to 0-1, thus will lose the absolute density information. 

\noindent\textbf{Coordinate regression.} Coordinate regression
methods can be divided into two-stage methods~\cite{lscnet,wan2021generalized} and end-to-end methods~\cite{gao2020learning,ppnet,LOWB,laradji2018blobs}. Two-stage methods usually predict more objects and then refine them by a dataset prior information such as a pseudo bounding box.

LSC-CNN~\cite{lscnet} first estimates pseudo-bounding boxes from the training sets using minimal neighbor distance. The pseudo-bounding boxes are used to perform NMS to filter out redundant predictions.

Generalized loss~\cite{wan2021generalized} applies radius rather than a pseudo-bounding box, which is easier to estimate and more suitable for the localization of human heads. However, the bounding box size or radius length is hard to estimate but sensitive to the final performance in this method. 

The end-to-end method can predict coordinates without needing other processes.
Javier Ribera~\cite{LOWB} applies weighted Hausdorff distance as set distance metrics and extends this distance into an average form to make it robust to outlier points. The P2PNet~\cite{ppnet} uses Hungarian matching to obtain the optimal one-to-one match between the target and the predicted point and only learns the matched GT points. Both of those methods face a problem: determining the threshold for positive, as it is evident that a fixed threshold is poor for crowd scenes. IIM~\cite{gao2020learning} and LSC-FCN~\cite{laradji2018blobs} use a binarize module to learn a threshold map, but a time-consuming post-process is needed to transfer the medium map to coordinates.

\section{Method}
\begin{figure*}
    \centering
    \includegraphics[width=0.8\textwidth]{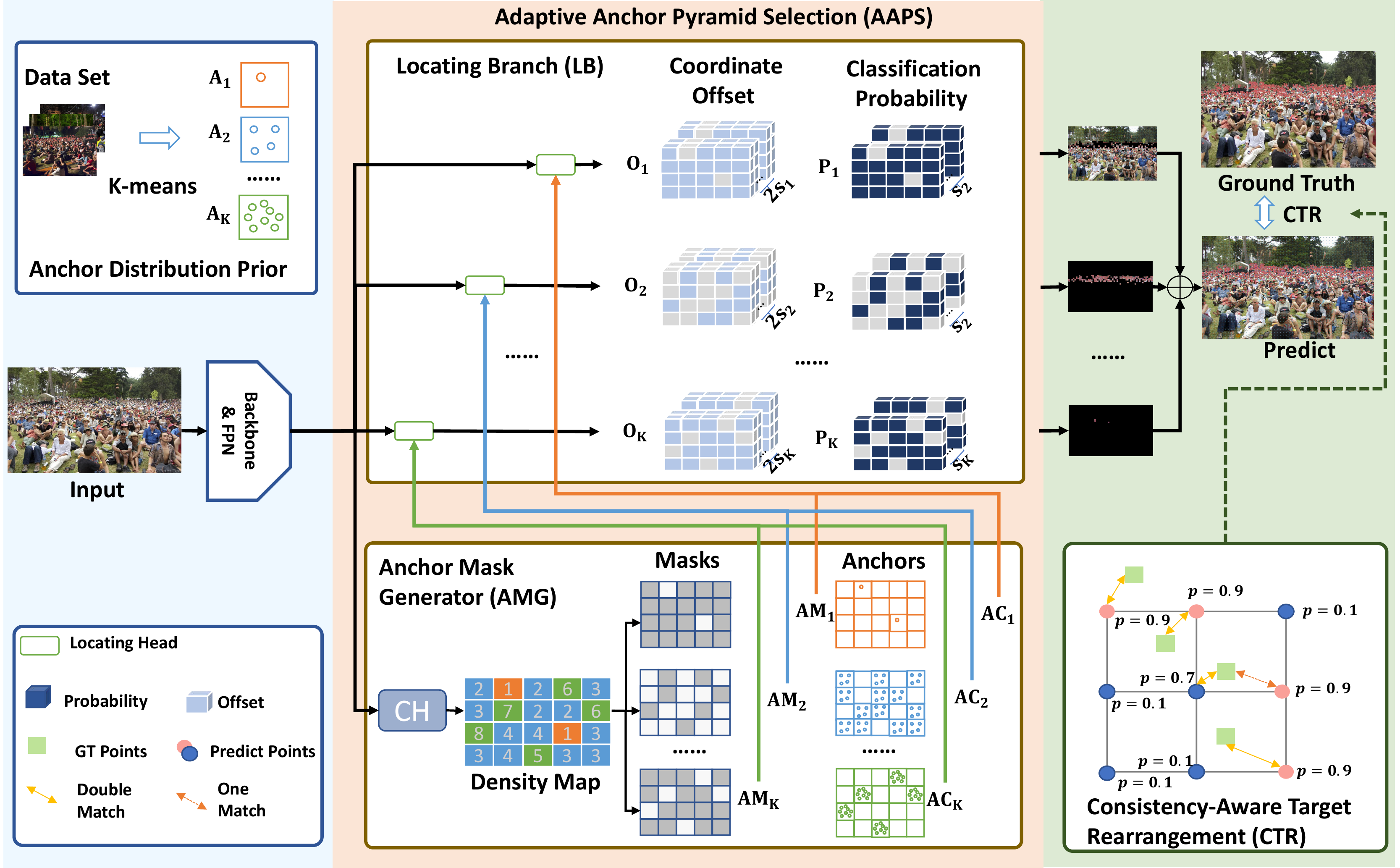}
    \caption{The architecture of the proposed method, which contains two main components:  an adaptive anchor pyramid selection strategy (AAPS, Sec~\ref{sec:aaps})  and a consistency-aware target rearrangement (CTR, Sec~\ref{sec:ctr}). 
    AAPS consists of a Locating Branch (LB) and an Anchor Mask generator (AMG). The input image is evenly divided into grids and LB predicts target locations using different anchor densities in each grid. 
    Then, AMG predicts a target density  for each grid, which is used to guide the selection of anchor density. Anchors at selected density are retained (others are discarded). All retained anchors serve as target candidates. 
    The proposed consistency-aware target reassignment (CTR) module then finds the best matching between retained anchors and ground truth annotations. Losses of matched anchors are used for gradient back-propagation.
    }
    \label{fig:pipeline}
\end{figure*}

As shown in Figure.~\ref{fig:pipeline}, besides the backbone, our model consists of two main components:
Adaptive Anchor Pyramid Selection(AAPS) and Consistency-aware Target Rearrangement (CTR).
Taking image features as input, AAPS first adopts a counting head (CH) to obtain density maps, which are used to generate anchor masks (AM). Combining the anchor distribution prior learned from the training set and anchor mask, unevenly distributed anchors are generated to handle various crowd densities efficiently. Then, the anchors, as well as anchor masks and FPN features, are fed into locating heads used to regress points, coordinate offsets, and probability. Finally, consistency-aware target rearrangement (CTR) is used to evaluate the loss between predicted candidates and ground truth points. As mentioned in ~\cite{savner2022crowdformer}, long-range feature interaction is essential, so we adopt ConvNext-S~~\cite{liu2022convnet} as the backbone.
\subsection{Adaptive Anchor Pyramid Selection}\label{sec:aaps}

Our \textbf{AAPS} consists of an anchor mask generator (AMG) and locating branch (LB). AMG predicts density maps using FPN features and generates anchor masks ($AM$) according to each feature grid's predicted density values $d$. The anchor masks are used to select anchor priors $A_i$ to determine the anchor density in each image region adaptively. Using the generated anchors, anchor masks, and features from FPN as input, LB predicts the coordinates and probability for objects. We denote the grid size by $h*w$, and for a $H*W$ image, there are $N_g$ grids, where $N_g=(H*W)/(n*w)$.

Unlike previous methods~\cite{ppnet,defcn,sun2021makes} that apply fixed numbers of anchors (or proposals) in each grid, AAPS dynamically generates anchors based on the density map to fit the crowd rate variety. 
Inspired by FPN, we design anchor sets of different densities and select an adaptive anchor set for each grid according to its density information.

Specifically, there are $K$ anchor prior sets represented as $\{A_1, A_2, \cdots, A_{K}\}$, and anchor set $A_i$ contains $s_i$ anchors.  
If the predicted density value in grid $(u,v)$ is in $[s_i,s_{i+1})$, we select anchor sets $A_i$ for this grid.
As a result, we generate $K$ anchor masks $\{AM_1,AM_2,\cdots,AM_{K}\}$, the shape of which is $(H*W)/(n*w)$, for coordinates prediction:
\begin{equation}
    AM_i(u,v)=\left\{
\begin{array}{cc}
     1 & s_{i}<\mathbf{D}(u,v)\leq s_{i+1} \\
     0 & \mbox{others}
\end{array}
\right.
\label{eq:am}
\end{equation}
where $\mathbf{D}(u,v)$ denotes the predicted number of objects among grid $(u,v)$.

The relative positions of anchors within a grid are learned by performing K-means on ground truth point annotations in the training dataset to achieve better priors for the network. Since we want to learn anchor priors for the grid, we crop training images into the grid size (16x16) and run K-means on the ground truth points in grids with Euclidean distance of points as the distance metric. 
To discover priors for anchors of different densities, we conduct K-means $K$ times by setting cluster number $\in \{s_1,s_2,\cdots,s_{K}\}$. The cluster centers denoted as $\{ac_i^1,ac_i^2,...ac_i^{s_i}\}, i\in\{1,\cdots, K\}$ are used as positions of anchors in anchor sets $AC_i$.
\begin{equation}
    AC_i(u,v)=\left\{
\begin{array}{cc}
     \{ac_i^1,ac_i^2,...ac_i^{s_i}\} & AM_i(u,v)==1 \\
     \empty & \mbox{others}
\end{array}
\right.
\label{eq:ac}
\end{equation}

To predict points based on anchor sets of different densities, we design $K$ heads in our locating branch. The $i^{th}$ head ($\mathbf{LH}_i(\cdot)$) takes the $i^{th}$ pyramid level features from FPN, $AM_i$ and $A_i$ as input, and output the probability for objects ($\mathcal{P}_i$) and its relative coordinates offset ($\mathcal{O}_i$) to the corresponding anchor points, 
\begin{equation}
    \mathcal{P}_i,\mathcal{O}_i=\mathbf{LH}_i(\mathbf{FPN}_i*AM_i, A_i)
\end{equation}
where $\mathbf{FPN}_i$ denotes the $i^{th}$ level FPN features. $\mathcal{O}_i$ includes two channels, which are the relative offset to the anchor points $AC_i$.
The absolute coordinates $(\mathbf{x},\mathbf{y})$ prediction of anchor $A_i^j$ at grid $(u,v)$ in the image is calculated by: 
\begin{equation}
    \mathbf{x}_{u,v,i,j}=u+ac_i^j+\textbf{Sigmoid}(\mathcal{O}_i(u,v,0))H
    \label{eq:ptx}
\end{equation}
\begin{equation}
    \mathbf{y}_{u,v,i,j}=v+ac_i^j+\textbf{Sigmoid}(\mathcal{O}_i(u,v,1))W
    \label{eq:pty}
\end{equation}

During the inference time, in grid $(u,v)$, the points with top $\mathbf{D}(u,v)$ classification probability is denoted as the positive prediction. 

As the generation of anchor masks depends on the density maps, unlike popular density map based crowd counting methods applying smoothing density map, which is more accessible for learning but outputs decimal values, our counting head is designed to predict integer numbers.
To this end, we select point maps rather than density maps as targets, which are too sharp to learn directly by convolutions~\cite{MCNN}. We observe that the popular loss for density maps are pixel-wised, such as Mean-Squared Error (MSE) or Mean-Absolute Error (MAE), which is too strict for tiny localization shift error. Thus when using point maps as learning targets, it can only determine what is better prediction if they are strictly matched to the point map.
As there is no precisely defined center on various appearances of human heads, the points can be annotated in a region nearby heads. So it is hard for the network to distinguish one pixel as positive and simultaneously predict its neighbor points as negative. Motivated by this, we propose to use a CasCade Region Loss (CCRL) rather than pixel-wise loss to learn a point map, which can be defined by,
\begin{equation}
	\mathcal{L}_{cnt}^{cas} = \sum_{r=0}^{t} \sum_{R\in T_i}
	\underbrace{\dfrac{
	    2^{r+1}
	}{H*W}}_{\text{size reduction}}
	\underbrace{\dfrac{
	    \mathbf{e}^{\mathcal{L}_1^{r-1}(R')}
	}{\mathbf{e}^{\sum_{R'\in T_{i-1}} \hat{\mathcal{L}}_1^{r-1}(R')}
	}
	}_{reweight}\hat{\mathcal{L}}_1^{r}(R)
	\label{eq:countingl}
\end{equation}
where $\hat{\mathcal{L}}_1^{r}(R)$ denotes the counting loss in region $R$ with size $2^r$, and can be calculated by 
\begin{equation}
	\hat{\mathcal{L}}_1^{r}(R) = \dfrac{1}{B}\sum_{i=1}^B |\mathcal{D}^i_R-\hat{\mathbf{D}^i}_R|
	\label{coutingl}
\end{equation}
where $B$ is the batch size, and the ground truth of object numbers in the region $R$ in cascade level $i$ are denoted as $\hat{\mathbf{D}^i}_R$.
The loss defined by Eq.~\eqref{eq:countingl} models the counting error from low resolution to high resolution and thus reduces the impact of localization shift errors. 
In Eq.~\eqref{eq:countingl}, $\dfrac{2^{r+1}}{H*W}$ is used to make loss reduction by the number of regions. If the loss calculated on the low-resolution region is small, then the error in sub-regions within this region is more likely to be a localization shift error. Therefore, we apply $\dfrac{\mathbf{e}^{\mathcal{L}_1^{r-1}}}{\mathbf{e}^{\sum_{R\in T_{i-1}} \hat{\mathcal{L}}_1^{r-1}}}$ to decrease the weight of $\hat{\mathcal{L}}_1^{r}(R)$. 
 
As the decimal number in $\mathbf{D}$ will be quantified to a round number, to reduce the impact of rounding operation on the accuracy, we add a regular term to encourage the output to be close to a round number by:
 \begin{equation}
     \mathcal{L}_{cnt}^{reg} = \dfrac{1}{B}\sum |\left[\mathbf{D}(u,v)\right]-\mathcal{D}(u,v))|
 \end{equation}
where $\left[\mathbf{D}(u,v))\right]$ denotes the round number of the predicted density value in grid $(u,v)$. 
The cascade region loss is defined as: 
\begin{equation}
    \mathcal{L}_{cnt}=\mathcal{L}_{cnt}^{cas}+\mathcal{L}_{cnt}^{reg}.
\end{equation}

\subsection{Consistency-aware Target Rearrangement}\label{sec:ctr}
\begin{figure}
    \centering
    \includegraphics[width=0.4\textwidth]{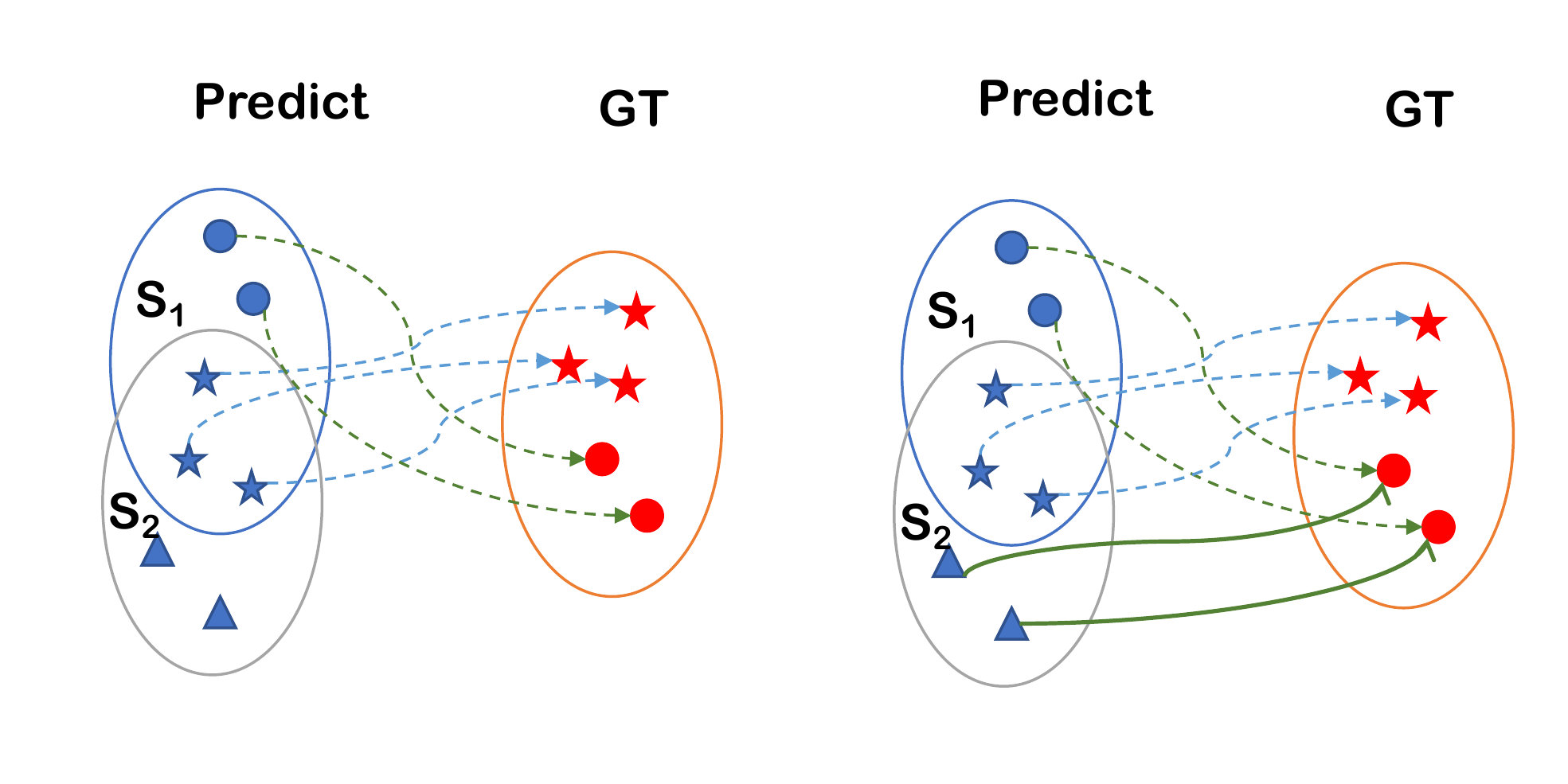}
    \caption{Traditional One-to-One match strategy (Left) and the proposed CTR match strategy (right). Except for candidates selected by $\mathcal{L}_{loc}$(dashed line), we add more candidates (denoted as blue triangles) with high classification probability and associate them with GT annotations using $\mathcal{L}_{dist}$ as metric (solid line).}
    \label{fig:CAM}
\end{figure}

Combing all the candidates predicted by our $K$ locating heads, we obtain the candidates points $S=\cup_{i=1}^K \{\mathcal{P}_i,\mathcal{O}_i\}$. For simplicity, we denote $S=\{t^i=(x^i,y^i,p^i)|t_1,t_2,\cdots,t_{N_s}\}$, where $p^i$ is the predicted probability.
The number of these candidates $N_S$ is: 
\begin{equation}
    N_S=\sum_{u=0}^{H/h}\sum_{v=0}^{W/w}\sum_{i=1}^{K}AM_i(u,v)s_i
\end{equation}
During training, in order to calculate training loss, the predicted candidate points need to be assigned to ground truth coordinates and classification scores.
The training loss for locating heads consists of classification and distance losses:
\begin{equation}
    \mathcal{L}_{loc}(S,G,\Omega) =\sum_{t^i \in S}\mathcal{L}_{cls}(p^i,\hat{p}^i)+\mathcal{L}_{dist}(x^i,y^i,\hat{x}^i,\hat{y}^i)
    \label{eq:locate}
\end{equation}
where $G$ denotes the sets of ground truth points, $\Omega$ denotes a match between ground truth points and predicts candidates. $t^i$ is matched to ground truth point $\hat{t}^i=\Omega(G,t^i)$ under $\Omega$, and $\hat{p}^i=1$ denotes that $t^i$ is assigned to a ground truth point, the position of which is $\hat{x}^i,\hat{y}^i$. We apply focal loss~\cite{lin2017focal} as the classification loss to balance the weight of foreground and background and set parameter $\lambda=2,\alpha=0.25$:
\begin{equation}
    \mathcal{L}_{cls}(t^i,\hat{t}^i)=-\alpha p^i(1-\hat{p}^i)\lambda \log(\hat{p}^i)+\hat{p}^i\lambda(1-p^i)\log(1-\hat{p}^i)
\label{focal}
\end{equation}
and apply L2 loss as the distance loss:
$\mathcal{L}_{dist}(t^i,\hat{t}^i)=\sqrt{(x^i-\hat{x}^i)^2+(y^i-\hat{y}^i)^2}$.
First, like the assignment rule mentioned in ~\cite{ppnet,peize2020onenet,defcn}, we apply training loss as the cost function to find the optimal match using Hungarian Match. 
We define $\Omega_1$ as the optimal match between $S$ and $G$:
\begin{equation}
    \Omega_1 = \mathop{\arg\min}\limits_{\Omega} \mathcal{L}_{loc}(S,G,\Omega)
\end{equation}
We refereed the matched candidate points determined by $\Omega_1$ as dual matched candidates $S_1$, because the cost metric considers both distance loss and classification loss. Formally, $S_1$ is defined as,
\begin{equation}
    t^i \in S_1 \gets \hat{p}^i=1
\end{equation}
Applying Eq.~\eqref{focal} as the cost metric for the optimal match, we select a set of candidates $S_2$ from $S$. Both of the size of $S_1$ and $S_2$ are $\sum\hat{\mathbf{D}}$. We introduce $S'\gets S_2-(S_2 \cap S_1)$. The candidates in $S'$ have high classification probability value and high $\mathcal{L}_{loc}$. 
These candidates will be selected during the inference time, although they are not assigned to ground truth points. In other words, candidates in $S'$ are not optimized during training.

\begin{algorithm}
\small
\caption{Matching strategy for proxy candidates}\label{alg:cap}
\begin{algorithmic}
\Require $S, G$
\Ensure $\Omega_2$
\State $\Omega_1 \gets \mathop{\arg\min}\limits_{\Omega} \mathcal{L}_{loc}(S,G,\Omega)$
\State $S_1,S_2,S',G'\gets \{\},\{\},\{\},\{\}$
\State $p_{high}\gets Topk(\mathcal{P},\sum\hat{\mathbf{D}})$
\For{$t\in S$} 
    \State $x,y,p \gets t$
    \State $\hat{t} \gets \Omega_1(G,t)$
    \State $\hat{x},\hat{y},\hat{p}\gets \hat{t}$
    \If{$\hat{p}==1$}
        \State $Add(S1,t)$
    \EndIf
    \If{$p > p_{high}$}
        \State $Add(S_2,t)$
    \EndIf
\EndFor
\State $S'=S_2-(S_1\cap S_2)$
\State $p_{low}\gets Topk(\mathcal{P},S'_{size})$
\For{$t \in S_1$}
    \If{$\hat{p}==1$}
        \State $Add(G',\hat{t})$
    \EndIf
\EndFor
\State $\Omega_2 \gets \text{argmin}_{\Omega} \mathcal{L}_{dist}(S',G',\Omega)$
\end{algorithmic}
\label{ag:s2selection}
\end{algorithm}

As shown in Figure.~\ref{fig:CAM}, we design a reassignment rule to reduce the inconsistency between training and inference. Under match $\Omega$, some ground truth points are assigned to prediction in $S_1$ with a low classification score. We denote those ground truth points $G'$. 

Those predictions are easy to be distinguished as negative predictions in inference time. The size of $G'$ is limited to the same as $S'$ by ranking the classification score to its corresponding prediction in $S_1$. We assign $S'$ as proxy candidates for $G'$. The metric apply to match $S'$ and $G'$ distance cost $\mathcal{L}_{dist}$. Those proxy candidates are also optimized with $\mathcal{L}_{dist}$ to encourage them to predict $G'$. It should be noted that those candidates are still negative samples to satisfy the one-to-one label assignment strategy. The details of matching strategy $\Omega_2$ for $S'$ and $G$ are demonstrated in Procedure~\ref{ag:s2selection}. Then the loss function can be defined by:
\begin{equation}
\begin{aligned}eee
    &\mathcal{L}_{loc}(S,G) =\sum_{t^i\in S}\mathcal{L}_{cls}(p^i,\hat{p}^i)+\\
    &\sum_{t^i\in S_1}\mathcal{L}_{dist}(x^i,y^i,\hat{x}^i,\hat{y}^i)+\sum_{t^i \in S'}\mathcal{L}_{dist}(x^i,y^i,\hat{x}^i,\hat{y}^i)
\end{aligned}
\label{eq:locate2}
\end{equation}

\section{Experimental Results}
We evaluate our approach on three datasets: ShanghaiTech, JHU-CROWD++, and UCF-QNRF. In this section, we first evaluate and compare our method with the previous state-of-the-art approaches to these datasets. Then we present ablation study results on JHU-CROWD++.
\begin{table*}[]
    \centering
    \footnotesize
    \rowcolors{1}{}{lightgray}
\begin{tabular}{cccccccccccccc}
        \hline
        \multirow{2}*{Methods} & \multirow{2}*{Backbone(flops)} &\multicolumn{3}{ c }{STA} & \multicolumn{3}{ c }{STB} & \multicolumn{3}{ c }{UCF-QNRF} & \multicolumn{3}{ c }{JHU-CROWD++}\\\cmidrule{3-14}
        &&F1&P&R&F1&P&R&F1&P&R&F1&P&R\\\hline
        TinyFaces\cite{tinyface} & ResNet-101 & 57.3&43.1& \textbf{85.5}&71.1& 64.7&79.0& 49.4 & 36.3& 77.3&-&-&-\\
        RAZLoc\cite{razloc} & VGG-16 & 69.2&61.3&79.5&68.0&60.0&78.3&53.3&59.4&48.3& -&-&-\\
        LSC-CNN\cite{lscnet} &VGG-16 & 68.0 &69.6& 66.5& 71.2& 71.7& 70.6& 74.0& 74.6& 73.5&51.9&51.2&52.7\\
        IIM\cite{gao2020learning} & HRNet-48 & 73.9& 79.8& 68.7 &86.2& \textbf{90.7}& 82.1& 72.0& 79.3& 65.9&62.5& 74.0 &54.2\\
        FIDT\cite{liang2021focal} & HRNet-48 & 77.6& 77.0& 78.1 & 83.5&83.2&83.9&  78.9&\textbf{82.2}&75.9& 74.7&74.7&74.7\\
        DCST\cite{dsct} & Swin & 74.5& 77.2& 72.1& 86.0& 88.8& 83.3 &72.4& 77.1&68.2&-&-&-\\
        P2PNet\cite{ppnet} & VGG-16 & 74.7&73.0&76.5& 82.7&81.1&84.4 &74.6&72.6&76.7&-&-&-\\\hline
        Proposed Methods & VGG-16 & 77.1&76.5&77.8 & 84.9&84.2&85.6& 76.1& 75.1& 77.2& 78.8&80.1&77.5\\
        Proposed Methods & HRNet-48 & 80.3&79.8&80.7& 87.0&85.6&88.4& \textbf{79.3}&77.3&\textbf{81.3}& 80.7&82.0&79.4\\
        Proposed Methods & ConvNext-S &\textbf{82.8}& \textbf{83.5}&82.1 & \textbf{89.0}&87.2&\textbf{91.1}& 79.2&77.5&81.0& \textbf{82.5}&\textbf{84.7}&\textbf{80.5}\\\hline
    \end{tabular}
    \caption{Comparison to state-of-the-art methods on the four datasets, JHU-CROWD++,  UCF-QNRF, and ShanghaiTech Part A\&B (STA, STB). From left to right are F1-measure (F1), Precision (P), and Recall(R). }
    \label{tab:main_result}
\end{table*}
\subsection{Experimental Setups}
\noindent \textbf{Implementation Details.}
During the training, the data augmentations we apply include Color Jitter, Rand Scaling( from 0.5 to 1.5), Random crop( to 256x256), Random Rotate (from 0 to 45), and Horizontal Flip (the probability of which is 0.5). During the inference time, the image size is padded to an integer multiple of 64. For the large picture datasets, we limited the longest edge within 1920 as P2PNet\cite{ppnet} and kept the original aspect ratio.

The output stride is set to 16 for prediction. We set $K=3$, and for all mentioned datasets except QNRF,  $s_1, s_2$ and $s_3$ are set to 1, 4, and 8. For the UCF-QNRF dataset, $s_1, s_2$ and $s_3$ are set to 1, 8, and 16. We apply AdamW as the optimizer, and the learning rate is set to 1e-4 with a cosine scheduler. All the backbone is pretrained on ImageNet. The experiments are conducted on eight NVIDIA RTX 3090 (~ 24 GB GPU Memory). 
\begin{figure*}[!htb]
\centering
    \includegraphics[width=0.8\textwidth]{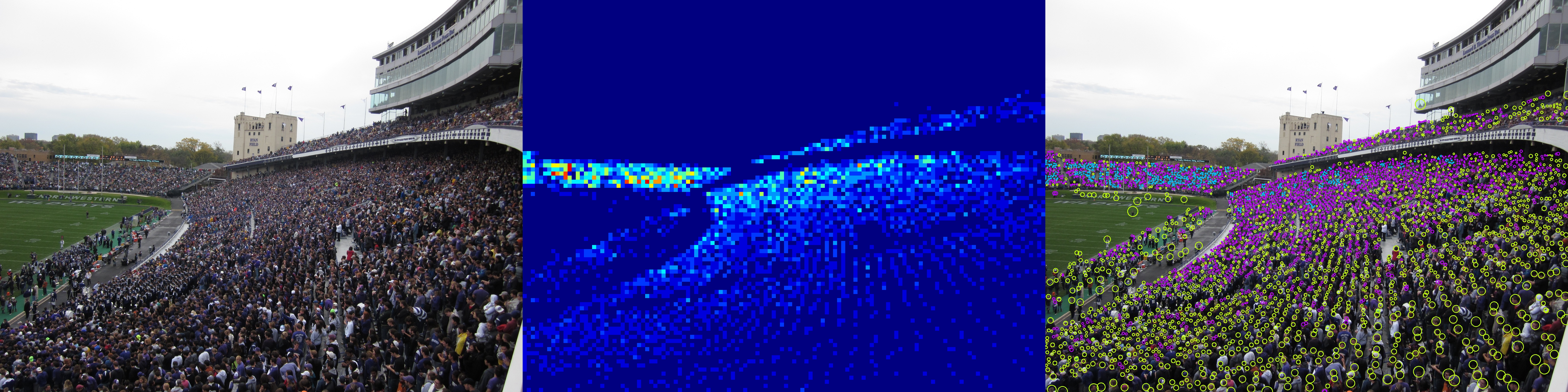}
\centering
    \includegraphics[width=0.8\textwidth]{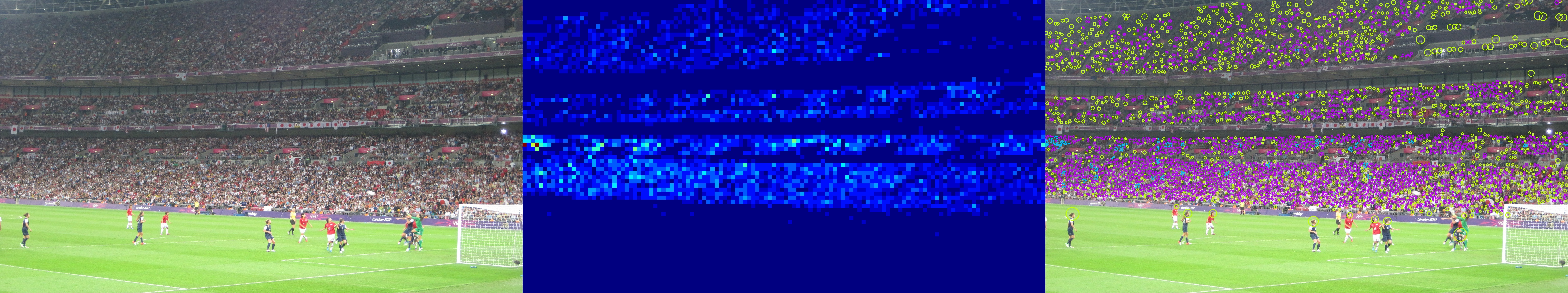}
\centering
\centering
    \includegraphics[width=0.8\textwidth]{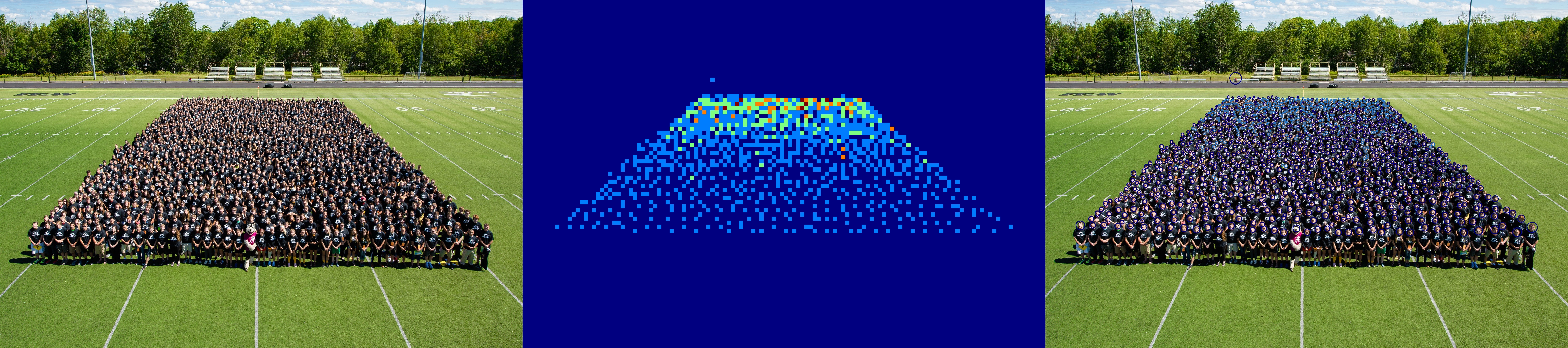}

\caption{Sample results of our methods on the JHU-CROWD++ test dataset (id 1795, 3178, 3560). 
Left to right: original images,  predicted density maps, and final outputs. 
In the density map, the color is deeper as the density is more significant. 
In the third column, the color of the circles denotes the level of the corresponding anchors at three levels, blue, yellow, and red, from the crowd to sparse. The circle sizes represent the estimated scale estimated by the method proposed in FIDT\cite{liang2021focal}. }.
 \label{fig:quan_results}
\end{figure*}

\noindent \textbf{Datasets and Evaluation Metrics}
ShanghaiTech Part A consists of web images that are very congested. Pictures in ShanghaiTech Part B are collected from streets and are relatively sparse. Pictures in the UCF-QNRF dataset are more challenging for their high resolution and much more absolute human counts (1525 images with human counts from 49 to 12865). The JHU-CROWD++ dataset is a more comprehensive dataset that covers multiple crowd rates and scenes (4250 images with human counts from 0 to 7286). Successful application of our methods on those datasets demonstrates its capacity to handle multiple crowd cases.

To evaluate the localization performance, we use Precision, Recall, and F1-measure (P, R, F1 for short) for evaluation where the positive rates are defined by $\sigma$ for the scale of objects~\cite{wang2020nwpu}. 
We set $\sigma_i$ for the JHU-CROWD++ dataset by 
$    \sigma_i=\sqrt{h_i^2+w_i^2} $
based on the height $h_i$ and width $w_i$ of the bound box annotation of points $i$.
For other datasets that have no bounding boxes annotations, we apply $\sigma=8$ for ShanghaiTech A\&B and an average metric for $\sigma$ from 1 to 100 for the UCF-QNRF dataset as FIDT\cite{liang2021focal}. 

To evaluate the counting performance, we apply Mean Absolute Error (MAE for short) and Mean Squared Error (MSE for short) as metrics.

\subsection{Evaluation Results}
In this section, we demonstrate the performance of our methods on previous on the four datasets. The main localization results compared with previous SOTAs (including density map based methods (IIM\cite{gao2020learning}, FIDT\cite{liang2021focal}, DCST\cite{dsct} ) and regression based methods (Tiny Faces\cite{tinyface}, RAZLoc\cite{razloc}, LSC-CNN\cite{lscnet}, P2PNet\cite{ppnet}) are shown in Table \ref{tab:main_result}. 

It should be noted that the backbone of previous methods is not unified. Some density map based methods (DCST\cite{dsct}, IIM\cite{gao2020learning}, FIDT\cite{liang2021focal}) adopted strong backbones (Swin or HRNet-48) for high-resolution predictions.
For a fair comparison, we test our method with different backbones, i.e., VGG-16, HRNet-48, and ConvNext-S.
It can be seen that under the same backbone, our methods achieve the highest F1-measure on all three datasets. Using VGG-16 as the backbone, our method increases the F1 measure by at least 2.1 (on the UCF-QNRF) compared with previous SOTA methods. Although TinyFaces\cite{tinyface} achieved the highest recall on STA, its precision is much lower than other methods. Using HRNet-48 as the backbone, the F1-measure on the largest dataset (JHU-CROWD++) is improved by 6 (from 74.7 to 84.7). 
Actually, the flops of HRNet-48\cite{hrnet} and Swin-B\cite{liu2021Swin} are 17.36G and 15.14G, respectively, whereas the flops of ConvNext-S\cite{liu2022convnet} is 8.69G (224x224).
Our method with ConvNext-S as backbone achieves higher F1-measure on ShanghaiTech Part (A\&B) and JHU-CROWD++. Because the resolution of images in UCF-QNRF is very high, HRNet-48 is more suitable for extracting image features. Our method with HRNet-48 obtains the best F1.

Adopting VGG-16 as the backbone, we also compare the counting performance of our method with SOTA crowd counting methods on those three datasets in Table \ref{tab:cnt_st}. It can be seen that, among all crowd localization methods, we achieved the top 2 counting performances comparable with crowd counting methods. It should also be noted that the thresholds for the best counting performance of FIDT\cite{lin2017focal} are not equal to that for the best localization performance. As a result, for FIDT, the number of predicted points is not equal to the predicted crowd number, while our method can ensure that the number of predicted points is consistent with the predicted crowd count.

In Figure.~\ref{fig:quan_results}, we show the visualization result on the JHU-CROWD++ test datasets. In the density map, the deeper the color, the larger the density. 
In the third column, the color of the circles in an image denotes the density level of their corresponding anchors. There are three levels of anchors denoted by blue, yellow, and red, from the crowd to sparse. It can be seen that our methods can apply a suitable adaptive base guided by the predicted density map. 

\subsection{Ablation Studies}

\begin{table}[]
    \centering
    \footnotesize
    \rowcolors{1}{}{lightgray}
    \begin{tabular}{ccccccc}
    \hline
        AAPS & CTR & F1 & P & R & MAE & MSE \\
        \checkmark&\checkmark&82.5(+6.3)&84.7&80.5& 64.2 & 258.7\\
        \checkmark & & 81.5(+5.3)&85.2&78.2&68.2&297.4\\
        &\checkmark& 78.7(+2.5)&81.1&76.4&62.7&255.0\\
        & & 76.2&78.6&73.9&66.2&299.6\\
        \hline
    \end{tabular}
    \caption{Ablation study of the main components.}
    \vspace{-2mm}
    \label{tab:ablation}
\end{table}

\begin{table}[]
    \centering
    \footnotesize
    \begin{tabular}{ccccccc}
    \hline
        Levels& $s_i$ & F1 & P & R & MAE & MSE \\\hline
         \rowcolor{lightgray}\cellcolor{white}&1&72.0&85.9&62.0&64.7&271.92\\
        &4&78.7&81.1&76.4&62.7&255.0\\
       \rowcolor{lightgray}\cellcolor{white} \multirow{-3}*{1}&8&76.0&65.1&91.2&65.6&274.2\\\hline
       & 1, 4, 8&82.5&84.7&80.5& 64.2 & 258.7\\
       \rowcolor{lightgray}\cellcolor{white} \multirow{-2}*{3}& 1, 2, 4&76.2&77.5&74.6&63.2&222.5\\\hline
        & 1,4,8,10 &82.1&84.3&80.0&63.8&235.7\\
        \rowcolor{lightgray}\cellcolor{white} & 1,4,8,12 & 81.8&83.7&80.0&62.8&240.3\\
        &1,4,8,14 & 81.1&83.7&78.7&64.1&254.4\\
     \rowcolor{lightgray}\cellcolor{white}\multirow{-5}*{4}& 1,4,8,16 & 81.2&84.5&78.9&65.0&224.0\\\hline
    \end{tabular}
    \caption{Ablation study of using different anchor pyramid levels.}
    \vspace{-2mm}
    \label{tab:apaag}
\end{table}

\begin{table*}[htb]
    \centering
    \footnotesize
    \rowcolors{1}{}{lightgray}
    \begin{tabular}{ccccccccc}
    \hline
    \multirow{2}*{Methods}& \multicolumn{2}{c}{STA}& \multicolumn{2}{c}{STB}& \multicolumn{2}{c}{JHU-CROWD++}& \multicolumn{2}{c}{UCF-QNRF}\\\cmidrule{2-9}
    &MAE&MSE&MAE&MSE &MAE&MSE&MAE&MSE\\\hline
         SAANet\cite{savner2022crowdformer} & 51.7& 80.1&6.1 &9.4& - &-& 79.3 &137.3\\
         ChfL\cite{Shu_2022_CVPR} & 57.5& 94.3& 6.9&11.0& 57.0 & 235.7& 80.3 &137.6\\\hline
         P2PNet\cite{ppnet}& \textbf{52.7}& \textbf{85.06}&\textbf{6.25}&9.9& -& -&85.32& 154.5\\
         LSC-CNN\cite{lscnet} & 66.4& 117.0& 8.1&12.7& -& -&120.5& 218.3\\
         FIDT\cite{liang2021focal} & 57.0 & 103.4& 6.9& 11.8&66.6 &253.6 &89.0& 153.5\\
         Ours & 57.4&101.2 & 6.6 & 10.7&64.2 & 258.7& \textbf{83.2} & \textbf{144.1}\\\hline
    \end{tabular}
    \caption{Counting results on ShanghaiTech Part A and B, JHU-CROWD++ and UCF-QNRF. The last four methods output coordinates.}
    \label{tab:cnt_st}
\end{table*}

\begin{table}[]
    \centering
    \footnotesize
    \rowcolors{1}{}{lightgray}
    \begin{tabular}{ccccccc}
    \hline
        K-means & count loss & F1 & P & R & MAE & MSE \\\hline
        \checkmark&&80.8&82.0&79.7&63.4&269.5\\
        &\checkmark&82.1&83.8&80.5&63.4&260.6\\\hline
        \checkmark&\checkmark&82.5&84.7&80.5& 64.2 & 258.7\\
        \hline
    \end{tabular}
    \caption{Ablations using different components of AAPS.}
    \label{tab:apaagcomp}
\end{table}

\begin{figure}
    \centering
    \includegraphics[width=0.4\textwidth]{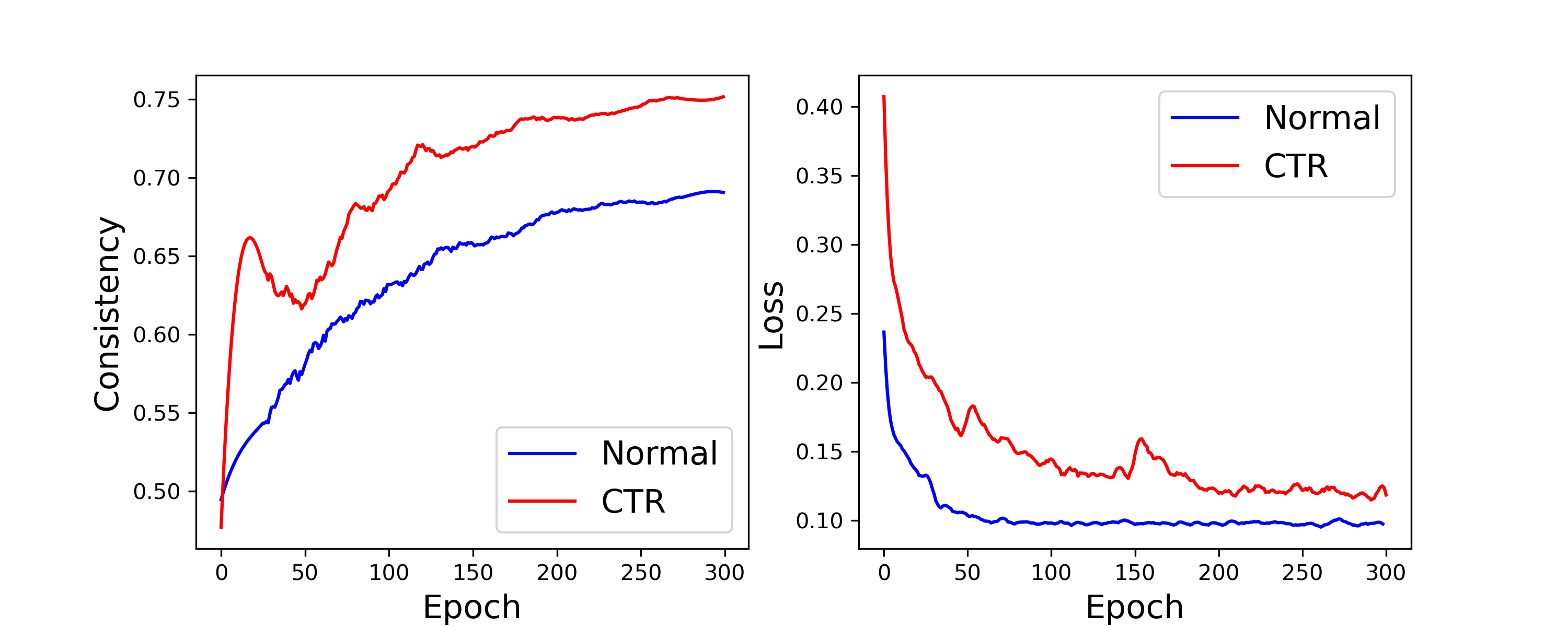}
    \caption{Consistency of the selected top K samples between training metric and counting metric.}
    \label{fig:consist}
\end{figure}

\noindent \textbf{Effectiveness of AAPS.} In Table \ref{tab:ablation}, we show the performance of with and without AAPS. For the experiment without AAPS we apply the 4 anchors in a $16\times16$ grid. Compared with using only one density of anchor, the anchor pyramid design achieves 5.6 F1-measure improvements and can get 3.7 F1-measure improvements with CTR. 

\noindent \textbf{Different Anchor Pyramid Level}. In our methods, we learn anchor priors of multiple density levels. Each density level is referred to as the anchor pyramid level (denoted as $K$ in our method). We conduct experiments with different $K$ to show the effect of anchor pyramid levels and report results in Table~\ref{tab:apaag}. As shown in Table \ref{tab:apaag}, when using only one pyramid level, when setting the anchor number to 4, the performance is the best. Compared to the first group and the second (or third) group, we can see that applying multiple anchor levels achieves better locating performance than using only one level. 
To verify whether more levels of pyramids are helpful, we design experiments by adding extra convolution on the FPN to get a stride 64 outputs and set its anchor numbers from 10 to 16. It can be seen that more levels of anchor do not helpful. It may be because there are too few ground truth points assigned to those anchors, making this branch not well trained.

\noindent \textbf{Effectiveness of Our Counting Loss}. To encourage round output of density map, we design region cascade loss defined by Eq~\eqref{eq:countingl}. To test its effectiveness, we replace our counting loss in Eq~\eqref{eq:countingl} with conventional MSE loss with $\sigma$ adaptively learned using a method proposed by \cite{MCNN}. The result reported in Table~\ref{tab:apaagcomp} shows that although using MSE loss can bring 0.8 MAE improvements (63.4 v.s. 64.2), the localization performance of MSE is worse than ours. The F1-measure decreases 1.7, demonstrating that the proposed cascade region loss in Eq~\eqref{eq:countingl} can improve the localization performance for AAPS with a tiny counting accuracy cost.

\noindent \textbf{K-means V.S. Uniform Distribution}. 
Besides anchor pyramids, AAPS also includes K-means to learn prior to datasets. Instead of K-means, we use anchors evenly distributed in the grid as \cite{ppnet}. The result is shown in Table~\ref{tab:apaagcomp}. The anchor prior obtained by K-means brings 0.4 F1-measure (82.1 v.s. 82.5) improvements.

\noindent \textbf{Effectiveness of CTR}. We replace our CTR with the matching strategy used in P2PNet\cite{ppnet}, and report the results in Table \ref{tab:ablation}. It can be seen that the proposed CTR reassignment rules improve the F1-measure by 2.5 points, and with the help of AAPS, the performance can keep improving by 1 point. To give an insight of the inconsistency problem in this task, we calculate the IOU of the set $S_1$ and set $S_2$ mentioned in Algorithm~\ref{ag:s2selection}. The result is shown in Figure.~\ref{fig:consist}. 
It can be seen that,
as the number of epochs increases, the training loss decreases, indicating that the model is convergent, but the IOU of the set $S_1$ and set $S_2$ is always less is far less than one. This means that there is a large inconsistency between the two sets.
When applying the proposed CTR, the consistency between these two sets are higher with about 0.1 IOU improvements. 

\section{Conclusions}
In this paper, we proposed an anchor pyramid network for crowd localization to predict the precise numbers of human coordinates. The main differences from existing methods are the Consistency-Aware Target Rearrangement (CTR) and Adaptive Anchor Generation(AAPS). CTR can reduce the ranking inconsistency between matched candidate points in the training and inference stage. The key component of CTR is to reassign ground truth points to the predicted candidates with high classification scores but not selected by the traditional label assignment rule. AAPS can adaptively determine the anchor density in each image region and provide anchors with higher predicting ability. The proposed simple framework achieves state-of-the-art localization performance and satisfactory counting accuracy.
\newpage
{\small
\bibliographystyle{ieee_fullname}
\bibliography{egbib}
}

\end{document}